\pdfoutput=1
\documentclass[runningheads]{llncs}
    \pdfoutput=1

\usepackage[misc,geometry]{ifsym}
\usepackage{bibunits}
    \usepackage[parfill]{parskip}
    \usepackage[utf8]{inputenc}
    \usepackage{amsmath,amssymb,amsfonts}%,amsthm}
    \usepackage{siunitx}
    \usepackage{booktabs}
    \usepackage{float}
    \usepackage{graphicx}

    \usepackage{caption, colortbl}
\usepackage{subcaption}
    \usepackage{tikz}
    \usetikzlibrary{shapes.geometric, arrows}
    \usepackage{algorithm, algpseudocode}
    \usepackage{pgfplots}
    \usepackage{cite, amssymb, multirow, array}
    
    \usepackage{todonotes, comment, wrapfig}
    \pgfplotsset{width=10cm,compat=1.9}
    
\newcommand{\bx}{\mathbf{x}}
\newcommand{\bmu}{\boldsymbol{\mu}}
\newcommand{\mX}{\mathcal{X}}
\newcommand{\Real}{\mathbb{R}}
\newcommand{\mF}{\mathcal{F}}
\newcommand{\mI}{\mathcal{I}}
\newcommand{\mGPns}{$\mathcal{G}\mathcal{P}$}
\newcommand{\mGP}{$\mathcal{G}\mathcal{P}~$}
\newcommand{\mD}{\mathcal{D}}
\newcommand{\mN}{\mathcal{N}}

\newcommand{\bt}{\mathbf{t}}

\newcommand{\prob}{p}
\DeclareMathOperator{\diag}{diag}
\newcommand{ \Given}{~|~}

\DeclareMathOperator{\entropy}{\mathbb{H}}
\newcommand{\trp}{^\top}\DeclareMathOperator*{\argmax}{\arg\!\max}
\newcolumntype{C}[1]{>{\centering\arraybackslash}m{#1}}
\DeclareMathOperator{\LatinHypercubeSampling}{LatinHypercubeSampling}

\usepackage{xcolor}

\newif\ifproofread

\graphicspath{{figures/}}
\date{}

\defaultbibliography{references}
\defaultbibliographystyle{unsrt}
    
\begin{document}
\proofreadtrue % change to remove colours for reviewers

\title{On Bayesian Search for the Feasible Space\\Under Computationally Expensive Constraints}

\author{Alma Rahat \Letter\inst{1}\orcidID{0000-0002-5023-1371} 
\and
Michael Wood \inst{2}\orcidID{0000-0002-4511-4445}
}
\authorrunning{Rahat \& Wood}
\titlerunning{On Bayesian Search for the Feasible Space}
\institute{Department of Computer Science, Swansea University, Swansea, UK.\\
\email{a.a.m.rahat@swansea.ac.uk}\\
\and
ACT Acoustics, Exeter, UK.\\
\email{m.wood@actacoustic.co.uk}
}
        
\maketitle

\begin{abstract}
We are often interested in identifying the feasible subset of a decision space
under multiple constraints to permit effective design exploration. If
determining feasibility required computationally expensive simulations, the cost
of exploration would be prohibitive. Bayesian search is data-efficient for such
problems: starting from a small dataset, the central concept is to use Bayesian
models of constraints with an acquisition function to locate promising solutions
that may improve predictions of feasibility when the dataset is augmented. At
the end of this sequential active learning approach with a limited number of
expensive evaluations, the models can accurately predict the feasibility of any
solution obviating the need for full simulations. In this paper, we propose a
novel acquisition function that combines the probability that a solution lies at
the boundary between feasible and infeasible spaces (representing exploitation)
and the entropy in predictions (representing exploration). Experiments confirmed
the efficacy of the proposed function.
\end{abstract}

%
%\begin{keywords}
%Feasible space, multiple constraints, computationally expensive problems, active learning, Bayesian search, surrogate-assisted learning, Gaussian processes.  
%\end{keywords}

\begin{bibunit}

\section{Introduction}

%In this paper, we want to construct a classifier that can   predict the feasibility of a solution when constraints are computationally expensive to evaluate. The aim is to use very few expensive evaluations 

In engineering applications, we are often interested in determining  the feasible %\footnote{The feasible space is sometimes referred to as the level set \cite{berkenkamp:safe}.}
design space for a given problem. This requires estimating a set of decision variables that
does not violate given conditions. This is a challenging task,
particularly if the constraints cannot be expressed analytically. In these
cases, computationally expensive simulations or physical experiments are required to explore the design space. For instance, in some nuclear power applications, keeping the neutron production
ratio below a critical level is essential for safe operation \cite{chevalier:fast}. This presents a significant design
challenge without analytical constraints. It is not practical to test each set of plant parameters by
simulation, since each evaluation of the simulator takes between $5$ and $30$
minutes. %An exhaustive search using Monte Carlo simulations is therefore not
%practical. %\cite{hsu:evaluating}.
Hence, a classifier that can accurately predict feasibility can allow operators to explore the design space rapidly before setting up the plant obviating the need for simulations. The challenge is then how to train this classifier with few expensive evaluations. 

%We propose a data-driven approach to such problems as surrogate-assisted methods are known to perform well
%for problems with a strict budget on the number of expensive evaluations
%\cite{chevalier:fast, jin:surrogate}.

%Apart from design exploration applications, the
%ability to determine the feasible space may also be useful in
%constrained global optimisation of problems for finding an initial feasible solution\cite{li:maxform} as well as
%integrating the probability of feasibility within the search
%\cite{picheny:sur-opt,bagheri:constraint}.

%In this paper, we focus on a surrogate-assisted sequential Bayesian search
%method. This method was inspired by the Efficient Global Optimisation (EGO) method \cite{jones:ego} (often referred to as Bayesian Optimisation).

In this context, surrogate-assisted Bayesian search method has been shown to be a data-efficient approach \cite{knudde:active}. This method starts with a small training set of independent parameters. These
parameters are expensively evaluated with a set of constraint functions. The resulting dataset is used to train a Bayesian regression
model (in this case, a Gaussian process, \mGPns) for each constraint
\cite{rasmussen:gpml}. Together, these models estimate the probability that a
given solution is feasible. In this way, the combination of models act as a
\textit{binary classifier}. 
%The challenge in creating this model is 
The idea is then to locate a candidate sample for evaluating expensively such that adding this sample to the training dataset would achieve the greatest improvement in the feasible space estimation. 
This candidate is located by maximising an \textit{acquisition function} (often referred to
as an infill criterion or a utility function). We keep adding new samples
until the budget on additional expensive evaluations is exhausted.
%
%To locate the next sample that should be evaluated expensively, and achieve the greatest improvement in the feasible space estimation when the training dataset is augmented, 
%We then expensively evaluate the next sample such that we achieve the greatest improvement in the feasible space estimation when the training dataset is augmented with the new evaluation. This candidate sample is located by maximising an \textit{acquisition function} (often referred to
%as an infill criterion or a utility function). We keep adding additional samples
%until the budget on additional expensive evaluations is exhausted.

We understand that using this method of Bayesian search for feasible region
identification and design exploration is new with Knudde \textit{et al.}
publishing the first acquisition function recently \cite{knudde:active}. This
function considers the loss of entropy of the posterior predictive distribution
from adding a new sample in the training dataset. With the aim of providing
alternative acquisition functions, the novel contributions
of this paper are:

%In this paper, we propose a new acquisition function, and investigate a range of alternatives.%, their
%approach only works when there is a single constraint for determining
%feasibility \cite{knudde:active}. Our approach address a scenario where there
%are multiple constraints. 
%Knudde \textit{et al.} has proposed an effective acquisition function that The novel contributions of our work are:
\begin{itemize}
\item A new acquisition function $\alpha_{PBE} (\cdot)$ based on
  the probability of a solution residing at the boundary (representing
  exploitation) and the entropy of predictive distribution (representing
  exploration). %We use this function to construct a classifier for predicting the
  %feasibility of a solution. %This function exhibits a high informedness using only a small
  %number ($11n$, where $n$ is the dimension of the decision space) of function
 % evaluations.
\item Adaptation of a range of alternative acquisition functions (that are originally used in reliability engineering for rapidly estimating the volume of the \textit{infeasible} space) for the purpose of data-efficient construction of a feasibility classifier.
\item A full investigation of these acquisition
  functions in a set of constrained problems. 
%  Bayesian search method for rapidly estimating the feasible space imposed by multiple computationally expensive constraints.
\end{itemize}

%In section \ref{sec:rel} we
%review related work from the reliability engineering literature.
In section \ref{sec:background}, we discuss necessary
concepts focusing on using \mGPns s to model constraints functions, and the
standard Bayesian search framework. Then we propose a range of acquisition
functions suitable for Bayesian search of the feasible space in section
\ref{sec:ac}. We present our results in section \ref{sec:exp}. Finally, we
finish with general conclusions in section \ref{sec:concl}.

\section{Background}
\label{sec:background}

Consider, a design vector $\bx$ in a design space $\mX \in \Real^n$.
Without loss of generality, a constrained problem with $L$
constraints can be defined as:
\begin{align}
G(\bx) = (g_1(\bx), \dots, g_L(\bx))\trp  \leq \bt = (t_1, \dots, t_L)\trp,
\end{align}
where, $g_l : \Real^n \rightarrow \Real$ is the $l$th constraint function with a threshold for feasibility $t_l$. 
To deal with equality constraints, we can add a small fixed constant $\epsilon$.
This converts the equation to an inequality constraint \cite{cec2006}.

The $l$th constraint function $g_l(\bx)$ generates a feasible space $\mF_l
\subseteq \mX$.
The infeasible set of solutions for this constraint is therefore
$\mI_l = \mX \setminus \mF_l$. The total infeasible set of solutions becomes
$\mI = \bigcup_{l=1}^L \mI_l$. 
If all constraints are considered, the feasible space
is at the intersection of all feasible sets: $\mF = \bigcap_{l=1}^L \mF_l$. 
%From a reliability engineering perspective, such a
%combination of constraints is considered as a parallel combination of multiple
%failure modes \cite{yang:system}.

%We can use a scalarisation approach as an alternative method for dealing with
%the multiplicity of constraints. This approach could encapsulate all constraints
%into a single function so that any violation of the scalarised constraint is
%equivalent to infeasibility \cite{li:maxform}:
% \begin{align}
% \label{eq:maxform}
%s(\bx) = \max_{l=1}^L \left(g_l (\bx)   - t_l\right) \leq  0.
% \end{align}
%
%Here, the response of $s:\Real^{n} \rightarrow \Real$ is only greater than $0$
%for a design vector resulting in an infeasible solution, iff at least one
%of the component $l$th constraints is violated ($g_l(\bx) > t_l$).
%From a reliability engineering perspective, \emph{mono-surrogate} approaches
%like this are known to be inferior \cite{yang:system}. We confirmed this to be
%true by a small test, and we exclude the approach from our investigation.

If constraint functions are cheap to evaluate, we can determine feasibility by
\emph{brute force} using Monte Carlo methods \cite{mori:mcs}. However, where
each constraint function evaluation $g_l(\bx)$ requires %an independently evaluated and
a computationally expensive simulation, this approach would be prohibitively slow.

\subsection{Modelling Constraints with Gaussian Processes}
\label{sec:gp}

Gaussian processes (\mGPns) are commonly used to construct surrogate models for
constraints $g_l(\bx)$. \mGPns s produce a Normal predictive distribution for
any arbitrary solution, producing a mean and standard deviation.

%\footnote{A comprehensive introduction may be found in \cite{rasmussen:gpml}.}. 
%The information provided by this
%predictive distribution can be used by the acquisition function to locate
%promising solutions.

In essence, a \mGP is a field of joint Gaussian distributions
\cite{rasmussen:gpml}. Consider a \mGP trained for $l$th
constraint function $g_l(\bx)$ on dataset $\mD_l = \{(\bx_m,
g_l(\bx_m))\}_{m = 1}^M$ evaluated at $M$ locations. The predictive probability for $g_l$ at $\bx$ is a
Gaussian distribution with mean $\mu_l(\bx)$ and variance $\sigma_l^2(\bx)$ is:
\begin{align}
\label{eq:gpmod}
% \prob (g_l  \Given  \bx, \mD_l, \theta_l) = \mN(\mu(\bx), \sigma^2(\bx)  \Given \bx, \mD_l, \theta_l).
\prob (g_l  \Given  \bx, \mD_l) = \mN(\mu_l(\bx), \sigma_l^2(\bx)  \Given \bx, \mD_l).
\end{align}

The efficacy of \mGPns s is conferred by a flexible kernel function.
We use a Matern $5/2$ kernel, as recommended for modelling
realistic functions \cite{snoek:practical}. We refer the reader to
\cite{rasmussen:gpml} for full documentation on how the \mGP is trained and
interrogated.

%where the mean and variance are
%\begin{align}
%\label{eq:mu}
%\mu_l(\bx) & = \bkappa(\bx, X) K^{-1} \bg_l \\
%\label{eq:sigma}
%\sigma_l^2(\bx)& =\kappa(\bx, \bx) - \bkappa(\bx, X)\trp K^{-1} \kappa(X, \bx).
%\end{align}
%
%Here $X \in \Real^{M \times n}$ is the matrix of design locations and
%$\bg_l \in \Real^M$ is the corresponding vector of the true function
%evaluations using $g_l(\cdot)$; thus $\mD_l = \{(X, \bg_l)\}$.
%The covariance matrix
%$K\in \Real^{M\times M}$ represents the covariance function
%$\kappa(\bx, \bx'; \theta_l)$ evaluated for each pair of observations and
%$\bkappa(\bx, X) \in \Real^M$ is the vector of covariances between $\bx$
%and each of the observations; $\theta_l$ denotes the kernel hyperparameters.

%We use the Matern $5/2$ kernel as the covariance function as this approach is
%recommended for modelling realistic functions \cite{snoek:practical}.
%To train a \mGP~model we estimating the hyperparameters $\theta_l$ by
%maximising the log likelihood of the data\footnote{We use the limited memory
%  BFGS algorithm with $10$ restarts to estimate the hyper-parameters
%  \cite{gpy}.}:
%\begin{align}
%\log \prob(\mD_l  \Given \theta_l) =& - \frac{1}{2} \log |K| - \frac{1}{2} \bg^\top K^{-1} \bs - \frac{M}{2} \log (2\pi).
%\end{align}
%
%In the following equations, we omit $\theta_l$ for simplicity as the elements in
%$\theta_l$ are set by maximum log likelihood estimates.

We train a model for each constraint independently. 
%For each $l$th constraint
%function and an arbitrary design vector $\bx$, we derive a posterior predictive
%distribution $p(g_l  \Given  \bx, \mD_l) = \mN(\mu_l(\bx), \sigma^2_l(\bx))$. Thus, 
Thus, the combined posterior predictive distribution across
all component models is a multi-variate Gaussian:
\begin{align}
\label{eq:prob-dist-G}
p(G  \Given \bx, \mD) = \mN(\bmu(\bx), \Sigma (\bx)) = \prod_{l=1}^L p(g_l  \Given \bx, \mD_l),
\end{align}
where, the training dataset is  $\mD = \{(\bx_m, g_1(\bx_m)), \dots,
g_L(\bx_m))\}_{m = 1}^M $, the mean prediction vector is $\bmu(\bx) =
(\mu_1(\bx), \dots, \mu_L(\bx))\trp$, and the predictive covariance matrix is
$\Sigma(\bx) = \diag(\sigma^2_1(\bx), \dots, \sigma^2_L(\bx))$. There are no
cross-covariances as each model is independent. %With this combined
%predictive distribution, we can compute the probability of feasibility.

\subsection{Classifying the Feasible Space}

\begin{figure}[b!]
\centering
\begin{tikzpicture}
\node[inner sep=0pt] (main) at (0,0)
    {\includegraphics[width=0.75\textwidth, trim={8mm 20mm 0 5mm}, clip]{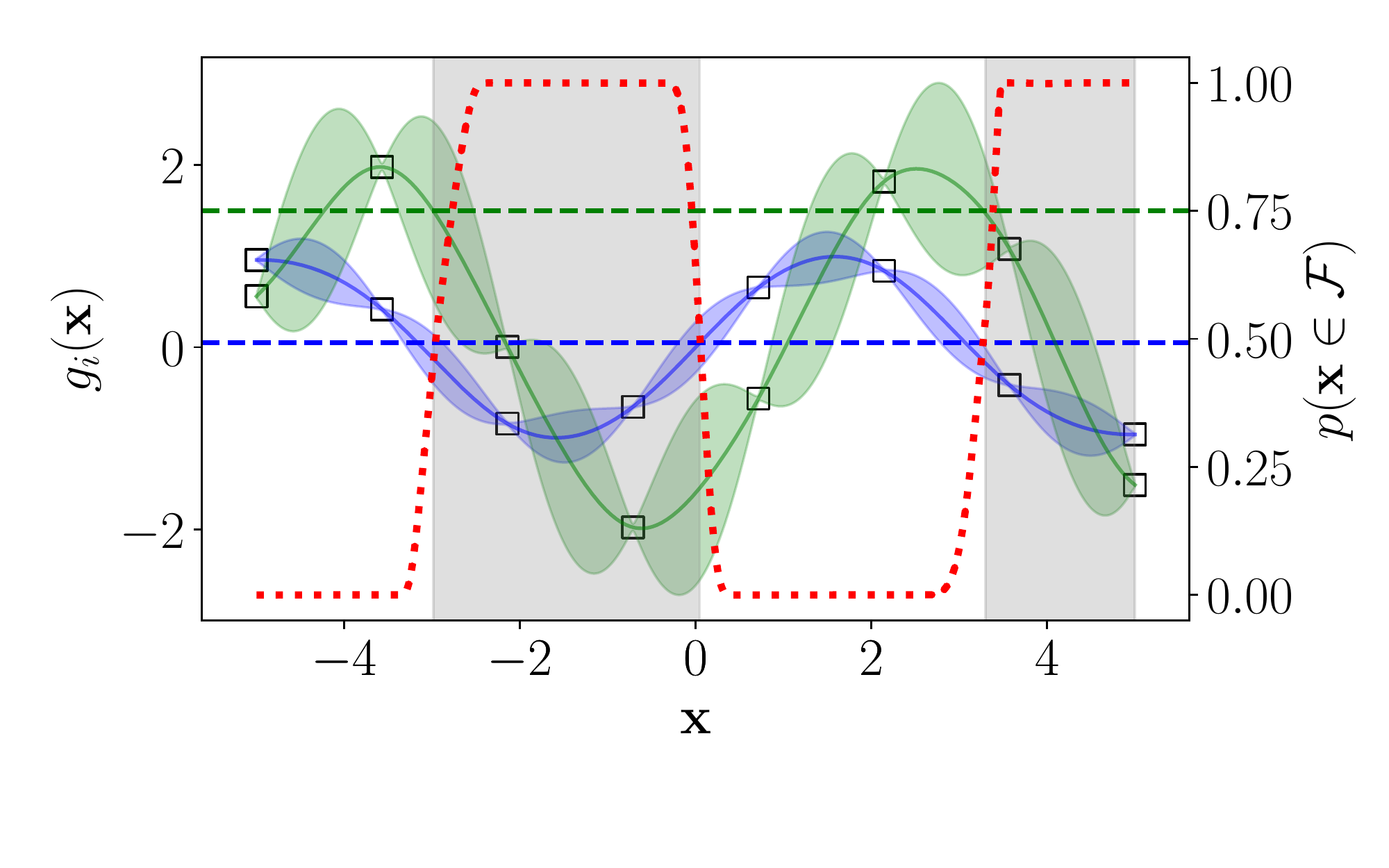}};
\node[inner sep=0pt] (labels) at (6,0.15)
    {\includegraphics[width=0.25\textwidth, trim={160mm 0mm 10mm 10mm }, clip]{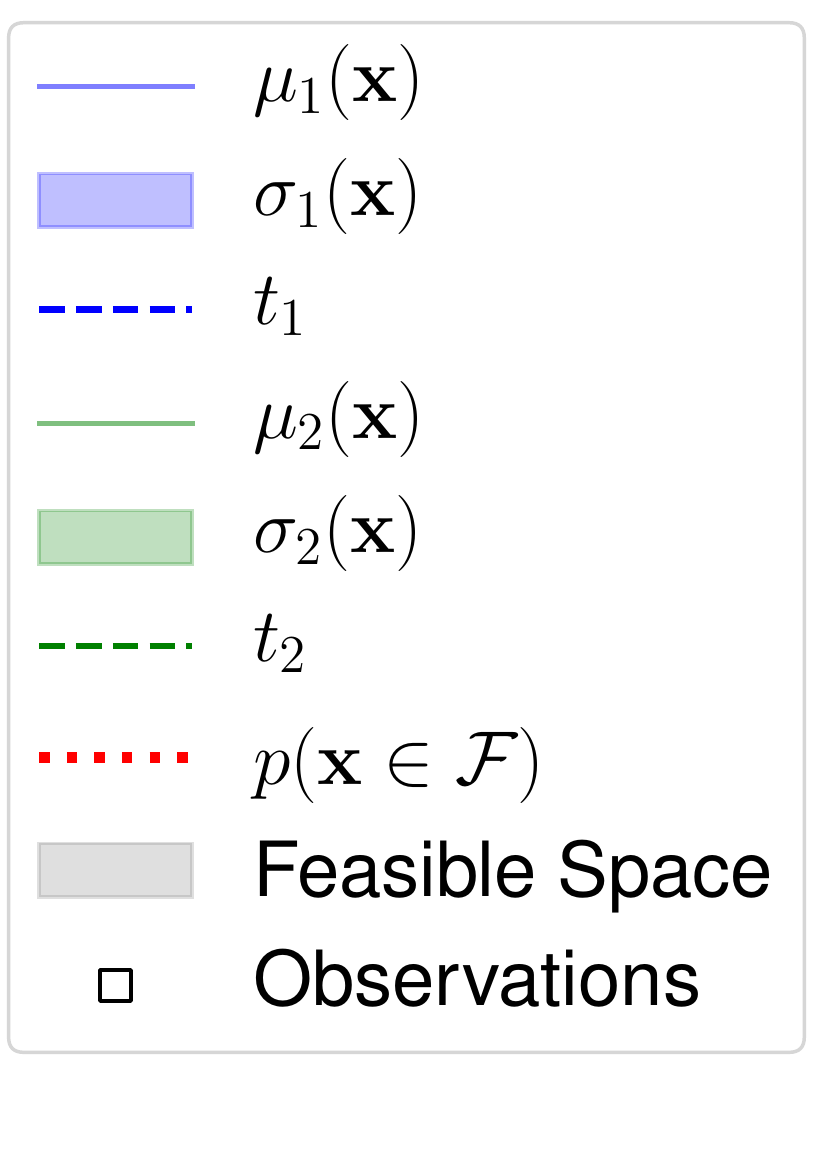}};
\end{tikzpicture}
\caption{
  Illustration of probability of feasibility (dotted red line) with two
  surrogate \mGPns s trained on observations at regular intervals (depicted by black
  squares). The grey shaded areas represent the true feasible space due to the
  threshold vector $\mathbf{t} = (t_1, t_2)^\top = (0.05, 1.5)^\top$ on
  constraint functions $g_1(\bx) = \sin (\bx)$ and $g_2(\bx) = 2 \sin(\bx - 1)$.
  The shaded areas around the mean predictions $\mu_{l}(\bx)$ (blue for
  $\mu_1(\bx)$ and green for $\mu_2(\bx)$) show the uncertainty estimations $2 \sigma_l(\bx)$ from respective \mGPns s. With only a few carefully selected
  observations, the models have well approximated the feasible space. In this paper, instead of sampling at regular intervals, our aim is to   
%  We then
  sequentially select the training data to construct the best possible
  classifier of feasibility with the smallest budget of expensive evaluations.}
\label{fig:prob-feas}
\end{figure}

Since the predictive distribution is Gaussian, we can
compute the probability of violation of each constraint.
For the $l$th constraint, the \textit{probability of feasibility} is
\cite{hughes:pdom, fieldsend:pdom,chevalier:kriginv}:
\begin{align}
p(\bx \in \mF_l) = p(~p(g_l  \Given  \bx, \mD_l) \leq t_l~) = \Phi( \tau_l),
\end{align}
where $\tau_l = \frac{t_l -  \mu_l(\bx)}{\sigma_l(\bx)}$ and $\Phi( \cdot)$ is the standard Gaussian cumulative distribution function. The overall probability of feasibility is therefore:
\begin{align}
\label{eq:prob-feas}
p(\bx \in \mF) = \prod_{l=1}^L p(\bx \in \mF_l) = \prod_{l=1}^L \Phi(\tau_l).%\left(\frac{t_l - \mu_l(\bx)}{\sigma_l(\bx)}\right).
\end{align}
Due to symmetry, the probability of infeasibility is $p(\bx\in\mI) = 1 -
p(\bx\in \mF)$. Using these probabilistic estimations, a decision vector $\bx$ is feasible \textit{iff} $p(\bx
\in \mF) > p(\bx \in \mI)$. Figure \ref{fig:prob-feas} illustrates the
predicted feasible spaces for two constraints modelled with two \mGPns s.

\subsection{Bayesian Search Framework}

Bayesian search is a surrogate-assisted active learning framework. This method
takes inspiration from Efficient Global Optimisation (EGO),
first proposed by Kushner \cite{kushner:ego} and later improved by Jones \textit{et al.} \cite{jones:ego}.
The framework can be used to minimise the mean squared error in the
sequential design of experiments, and is particularly useful where there are few observations \cite{sacks:design}. It has also been used to compute the volume of infeasible space \cite{bect:sequential, chevalier:fast, chevalier:kriginv}, and to \emph{locate} the feasible space under multiple constraints \cite{knudde:active}.

Bayesian search is a global search strategy. It sequentially samples the design
space to determine the boundary of the feasible space. The algorithm has two
stages: initial sampling, and sequential improvement.

During the first stage, we sample the parameters using a space filling design,
typically with Latin Hypercube sampling (LHS) \cite{mckay:lhs}. These parameters are
then evaluated by the true function. The LHS parameter samples and their
true-function output create the initial training set. Each design set is used to
create a set of models, one for each constraint, $\hat{G} = \{\hat{g}_1, \dots,
\hat{g}_L\}$.

For the sequential improvement phase, we can use $\hat{G}$ to locate promising
samples. 
For an arbitrary design vector $\mathbf{x}$, the function $\hat{G}$ provides a
multi-dimensional posterior distribution $p(G \Given \bx, \mD)$ with a mean
prediction (a vector) and uncertainty (a covariance matrix).
The predictive distribution permits a closed form calculation of probability.
We use this predictive distribution to estimate the likelihood that a
constraint function value will exceed a threshold. Since our goal is to minimise
the uncertainty around the threshold that bounds the infeasible space, we can
design our \emph{acquisition function} $\alpha(\bx, \hat{G}, \bt)$ accordingly.
The aim is to strike a balance between exploitation (through the probability of a solution residing at the boundary)
and global exploration (through the prediction uncertainty). In this way, the
acquisition function will drive the search towards the areas we are interested
in. The proposed acquisition functions are presented in section \ref{sec:ac}.

The most promising solution is defined where $\bx^* = \argmax_\bx \alpha(\bx, \hat{G},
\bt)$. We then determine $\bx^*$ expensively, and use the results to augment the
data and retrain $\hat{G}$.
We repeat this process until we exhaust the simulation budget.
When training is complete, we use $\hat{G}$ to estimate the feasible space.
For an arbitrary $\bx$ a probability of feasibility is returned using \eqref{eq:prob-feas}.
Algorithm \ref{alg:BS} summarises the method.

\begin{algorithm} [b!]
  \caption{Bayesian search framework. }
  \label{alg:BS}
  \textbf{Inputs}% Inputs section
  \begin{algorithmic}[]
    \State $M:$ Number of initial samples
    \State $T:$ Budget on expensive function evaluations
    \State $\bt:$ Threshold vector
   % \State $S:$ Boolean indicator; \emph{true} or \emph{false} to train a mono- or multi-surrogate model
  \end{algorithmic}
  \bigskip
  \textbf{Steps}
  \begin{algorithmic}[1]
	\State $X \gets \LatinHypercubeSampling(\mX)$ \Comment{\small{Generate initial samples}}
	\State $\Gamma \gets \{ G(\bx \in X)\}$ \Comment{\small{Expensively evaluate all initial samples}}
	\For {$i = M \rightarrow T$}
			\State $\hat{G} \gets $ Train\mGPns s$(X, \Gamma)$  \Comment{\small{Train a mono- or multi-surrogate model of constraints}}
			\State $\bx^* \gets \argmax_\bx \alpha(\bx, \hat{G}, \bt)$ \Comment{\small{Optimise acquisition function}}
		\State $X \gets X \cup \{\bx^*\}$ \Comment{Augment data set with $\bx^*$}
		\State $\Gamma \gets \Gamma \cup \{G(\bx^*)\}$ \Comment{Expensively evaluate $\bx^*$}
	\EndFor
	\State \Return $\hat{G}$ \Comment{Return trained models for feasibility classification using \eqref{eq:prob-feas}}
  \end{algorithmic}
\end{algorithm}

\section{Acquisition Functions}
\label{sec:ac}

For Bayesian search of the feasible space, the acquisition function's aim is to locate the boundary
between feasible and infeasible spaces: $\mF_l$ and $\mI_l$. A
solution based on $\hat{g}_l$ is often identified with the probability of being at the boundary: $p(\bx \in \mF_l) \prob (\bx \in \mI_l)$.
If we add the sample at this to the training set, the estimation of feasibility
with $\hat{g}_l$ is maximally improved. In this way, we achieve maximal
\emph{exploitation} of the latest knowledge of the model.

When data is limited, the uncertainty in predictions may be high, especially in
areas of the design space that have a sparse number of samples. We should, therefore,
promote the \emph{exploration} of these areas.

However, if we only prioritise sparsely populated areas, we may miss areas
near the threshold of interest. We therefore need to consider areas where 
\emph{both} the uncertainty and the boundary probability are high.
% This strikes a balance between myopic exploitation and global exploration, and aims to gain as much
% knowledge as possible about the boundary with every new addition to the training
% dataset.
We test both a range of existing and a novel acquisition
function for balancing the above requirements.

The first acquisition function for feasible region discovery was proposed by
Knudde \textit{et al.} \cite{knudde:active}. It is designed to maximise the loss
in entropy of the posterior distribution around the boundary for adding a
solution to the training dataset. For $l$th constraint, it can be expressed as:
\begin{align}
\label{eq:knudde}
\alpha_{K[l]}(\bx, \hat{g}_l, t_l) 
%= \entropy(\prob(g_l \Given \bx, \mD)) - \expected [\entropy (\prob(g_l \leq t_l \Given \bx, \mD))] 
= \frac{1}{2} \ln (2\pi e \sigma_l^2(\bx)) - \ln (\Phi(\tau_l)(1-\Phi(\tau_l))).
\end{align}

A detailed simplification of this equation is provided in the supplementary
materials.

Here, the first term is the predictive
entropy $\entropy(\prob(g_l \Given \bx, \mD))$ (representing areas of high uncertainty where exploration should be maximised).
The second term is the natural logarithm of the probability of being at the
boundary (representing exploitation of the knowledge of the boundary); c.f.
equation \eqref{eq:prob-feas}.

This function was originally designed for a single constraint. For multiple
constraints, they proposed to combine these component utilities across all
constraints as a sum:
\begin{align}
\alpha_K(\bx, \hat{G}, \bt) = \sum_{l=1}^L \alpha_{K[l]}(\bx, \hat{g}, t_l).
\end{align}

The summation formulation of the acquisition function permits a situation where
one of the components $\alpha_{K[l]}$ can dominate
the overall utility. For instance, if an arbitrary solution $\bx$ is expected to
gain significantly more information than other constraints, it may still
maximise the acquisition utility and thus get evaluated. This biases the
search towards maximal individual gain without allowing existing information to
influence over the value of the acquisition function.
As a result, the overall progress of this acquisition function may be slow.

%Some of the most popular acquisition functions can be adapted for Bayesian
%search. In this section, we describe how popular acquisition functions can be
%adapted. We also propose a new acquisition function.

%Rest of the acquisition functions 

We also \textbf{adapt} a range of alternative acquisition functions from the
field of reliability engineering. These acquisition functions were originally
developed for a single constraint, and were first used in an active learning
framework by Ranjan \textit{et al.} \cite{ranjan:single} and Bichon \textit{et
  al.} \cite{bichon:single}, and later popularised by Picheny \textit{et al.}
\cite{picheny:single, chevalier:fast} for computing the volume of the infeasible
space.

The most popular acquisition functions for \textbf{single} constraint are:
\begin{align}
\alpha_{T[l]}(\bx, \hat{g}_l, t_l) &=  \sigma(\bx) \phi(z),\\
\alpha_{B[l]}(\bx, \hat{g}_l, t_l) &= \sigma(\bx) [z^+ \Phi(z^+) + z^- \Phi(z^-) +
\phi(z^+) + \phi(z^-) - 2z\Phi(z) - 2\phi(z)],\\
\alpha_{R[l]}(\bx, \hat{g}_l, t_l) &= \sigma^2(\bx) [z^2 (\Phi(z^-) - \Phi(z^+)) +z^+ \phi(z^-) - z^-\phi(z^+)].
\end{align}
Here, $z = \frac{\mu_l(\bx) - t_l}{\sigma(\bx)}$, $z^+ = z  + 1$, $z^- = z-1$, and $\phi(\cdot)$ is the standard Gaussian probability density function. $\alpha_{T[l]}(\cdot)$ is the targeted mean squared error and was
defined by Picheny \textit{et al.} \cite{chevalier:kriginv}.
$\alpha_{B[l]}(\cdot)$ and $\alpha_{R[l]}(\cdot)$ are functions that compute a
form of average positive difference between uncertainty and the predictive
distance from the threshold, defined by Bichon \textit{et al.} \cite{bichon:single} and
Ranjan \textit{et al.} \cite{ranjan:single}. Further details of these can be found in
\cite{chevalier:kriginv, bect:sequential}.

A similar acquisition function proposed by Echard \textit{et al.} that can also be used \cite{echard:ak}.
This is written as \cite{echard:ak, yang:system}:
\begin{align}
\alpha_{E[l]}(\bx, \hat{g}_l, t_l) = -\frac{|\mu_l(\bx) - t_l|}{\sigma_l(\bx)}.
\end{align}

This is the negative of the probability of wrongly predicting feasibility.
Maximising this function finds solutions that reduce the
misclassification error.

%We refer the reader to Lv \textit{et al.} \cite{lv:entropy} and
%Sun \textit{et al.} \cite{sun:lif} for further work on single constraints.

To determine areas of system failure under multiple constraints,
a composite-criterion approach is commonly taken. This approach
calculates the acquisition function for each model, selecting a single model
based on the best individual mean prediction \cite{echard:ak, fauriat:ak, yang:system}.
We reformulated this approach into a generalised version appropriate for
Bayesian search (without requiring a large number of Monte Carlo samples):
\begin{align}
\label{eq:msurr-gen}
\alpha_{Y}(\bx, \hat{G}, \bt) = \alpha_{Y[k]}(\bx, \hat{g_k}, t_k) ~|~ k = \argmax_{l=1}^L (\mu_l(\bx) - t_l),
\end{align}
where, $\alpha_{Y[k]}(\bx, \hat{g_k}, t_k)$ is the acquisition function for $k$th constraint $g_k(\bx)$, with $Y \in \{T, B, R,
E\}$.

Using the acquisition function \eqref{eq:msurr-gen} improves the learning of the
individual boundaries between feasible and infeasible spaces for each constraint $g_k(\bx)$. 
However, this approach does not directly account
for the true boundary under multiple constraints. For multiple constraints, \emph{any}
violation is treated as infeasible, and since equation \eqref{eq:msurr-gen} may
sample infeasible space, it will likely introduce unnecessary redundancy. A
further weakness is that the model selection term $k = \argmax_{l=1}^L (\mu_l(\bx) -
t_l)$ does not consider prediction uncertainty. The result can therefore be
misleading. The scale of the function value in each constraint can also cause
problems, since the magnitude differences in $\mu_l(\bx) - t_l$ may be inverse
to relative importance. Our new acquisition function aims to solve these
shortcomings.

\subsection{\textbf{P}robability of Being at the \textbf{B}oundary and \textbf{E}ntropy (PBE)}

We have discussed how single-constraint acquisition functions can be
combined to create an acquisition function for multiple constraints. However,
since our aim is to find solutions with a high probability of being at the
boundary of the feasible space (\emph{exploitation}), whilst minimising the overall uncertainty in the models (\textit{exploration}), we combine these two objectives as a product.

The probability that a solution is at the boundary $\beta$ between the feasible
and infeasible spaces, given a multi-surrogate model $\hat{G}$, is:
 \begin{align}
p (\bx \in \beta) = p(\bx \in \mF) ~p(\bx \in \mI) = \prod_{l=1}^L \Phi(\tau_l) - \prod_{l=1}^L \Phi^2(\tau_l).
%\prod_{l=1}^L \Phi\left(\frac{t_l - \mu_l(\bx)}{\sigma_l(\bx)}\right) - \prod_{l=1}^L \Phi^2\left(\frac{t_l - \mu_l(\bx)}{\sigma_l(\bx)}\right).
\end{align}

If we maximise the probability over the design space, we will locate solutions
at the boundary, thereby exploiting the current knowledge.

To evaluate the overall uncertainty for a multi-surrogate model $\hat{G}$, we compute the differential entropy of a multi-variate Gaussian distribution:
\begin{align}
\entropy(\bx  \Given \hat{G}) 
= \frac{L}{2} \ln(2\pi e) + \frac{1}{2} \ln (|~\Sigma~|) \propto \ln\left(\prod_{l=1}^L \sigma^2_l(\bx)\right).
\end{align}

The extremes of the above equation identify the solutions with most overall uncertainty across the models. These extremes identify the most informative samples.

To maximise both quantities, we combine these two measures together as a
product. This is a somewhat greedy approach that ensures that a solution that improves all components is selected.  Our multi-surrogate acquisition function -- the
Probability of Boundary and Entropy (PBE) -- is defined as:
\begin{align}
\alpha_{PBE}(\bx, \hat{G}, \bt) = p (\bx \in \beta) ~ \entropy(\bx  \Given \hat{G}).
\end{align}

This function addresses the true boundary $\beta$, which is at the intersection of all component constraints' feasible spaces $\mF = \bigcap_{l=1}^L \mF_l$, directly. It is particularly
useful, since no explicit model selection is required. Further, since the
probability and entropy are being computed via an intra-constraint model (rather
than between constraints), we expect it to perform better for unscaled function responses.

\section{Experiments}
\label{sec:exp}

To test the performance of our approach, we used the test suite for constrained
single-objective optimisation problems from CEC2006 \cite{cec2006}.

We selected a range of problems with only non-linear inequality constraints (for
simplicity) and varying proportional volume of the feasible space between
$0.5\%$ and $45\%$ (Table \ref{tab:problems}). 
%We restricted the suite of test problems to those with a feasible space volume greater than $0.5\%$ (Table \ref{tab:problems}).
Note that we merely use these problems as example constrained problems for \textit{design exploration}, and we do not seek to locate the global optimum.
%starting from $n$ initial training samples where $n$ is the dimension of the decision space, we set our budget to only $11n$, i.e. we allow $10n$ further expensive evaluation from the initial set. 

% save original \intextsep
\newlength{\oldintextsep}
\setlength{\oldintextsep}{\intextsep}

\setlength\intextsep{0pt}
\begin{wraptable}[14]{R}{6.5cm}
\caption{A range of test problems with a feasible space volume $\rho =  \frac{|\mF| }{|\mF|  + |\mI|}\times 100\%\geq 0.5\%$ from the test suite defined in \cite{cec2006} and implemented in PyGMO \cite{pygmo}. Here, $n$ is the dimension of the decision space, and $L$ is the number of constraints. \label{tab:problems}}
\begin{tabular}{C{1.5cm}C{1.5cm}C{1.5cm}C{1.5cm}}
\hline
   ID &   $n$ &     $\rho(\%)$ &  $L$\\
\hline
    G4 &   5 & 26.9953 &     6 \\
    G8 &   2 &  0.8727 &     2 \\
    G9 &   7 &  0.5218 &     4 \\
   G19 &  15 & 33.4856 &     5 \\
   G24 &   2 & 44.2294 &     2 \\
\hline
\end{tabular}
\end{wraptable}

We ran each method starting from an initial sample size of $n$ where $n$ is the dimension of the decision space to avoid modelling an under-determined system. We set our budget on expensive evaluations to $11n$ to allow each method to gather $10n$ samples after initial sampling. This budget is less than the number of evaluations used by Knudde \textit{et al.} \cite{knudde:active} and Yang \textit{et al.} (in reliability engineering) \cite{yang:system} for reporting their results. 

% We ran each method $21$ times on each problem,
%starting from $n$ initial training samples and a total budget of $11n$
%evaluations

We ran each method on each problem $21$ times  \footnote{Python code for Bayesian search will be available at: \url{bitbucket.org/arahat/lod-2020}}. The initial evaluations are matched between acquisition functions, i.e. for each pair of problem and simulation run, the same initial design was used. The exception to this is the LHS with $11n$ samples.

Since the acquisition function landscape is (typically) multi-modal, we used
Bi-POP-CMA-ES to search the space, as it is known to solve multi-modal
problems effectively \cite{hansen:compare}. We set the maximum number of
evaluations of the acquisition function to $5000n$.

We use \textit{informedness} as a performance indicator for the classifier. The
informedness estimates the probability that a prediction is informed, compared
to a chance guess. We chose informedness (instead of F1 measure used in
\cite{knudde:active}) as it performs well for imbalanced class sizes, which are
common when comparing the sizes of feasible and infeasible spaces for real-world
constrained problems \cite{powers:evaluation, tharwat:classification}. To ensure
that we get an accurate estimation, we used $10000$ uniformly random samples
from the decision space for validation. We keep these validation sets constant
across different methods for a specific simulation number and a specific
problem.

We used the one-sided Wilcoxon Signed Rank test with Bonferroni correction
to test statistical equivalence to the best median performance, due to matched
samples. We identify the best method at the level of $p\leq 0.05$
\cite{knowles:tutorial}. We used Mann-Whitney-U test to compare the LHS to the
other methods (Table \ref{tab:res}). We provide box plots for performance
comparison in the supplementary sections.

\begin{table}[t!]
\centering
\caption{Performance of different acquisition functions in terms of median informedness ($\%$) and the median absolute deviation from the median (MAD). The red cells show the best median performance, while the blue cells depict the equivalent methods to the best. \label{tab:res}}
\begin{tabular}{cc C{1.4cm} C{1.4cm}C{1.4cm}C{1.4cm}m{1.4cm}C{1.4cm}C{1.4cm}}
\hline
                     &        & LHS & $\alpha_K$ & $\alpha_T$ & $\alpha_B$ & $\alpha_R$ & $\alpha_E$ & $\alpha_{PBE}$ \\
\hline
\multirow{2}{*}{G4}  & Median &  $99.83\%$   & $99.66\%$   &\cellcolor{blue!25}  $99.95\%$       & \cellcolor{blue!25}  $99.94\%$         &    $99.93\%$         &      \cellcolor{blue!25}  $99.95\%$     &    \cellcolor{red!25}   $99.99\%$        \\
                     & MAD    &   $7.9 \times 10^{-4}$  & $1.3 \times 10^{-3}$& \cellcolor{blue!25}    $2.7 \times 10^{-4}$       &     \cellcolor{blue!25}  $2.5 \times 10^{-4}$     &    $4.3 \times 10^{-4}$        &  \cellcolor{blue!25}         $2.1 \times 10^{-4}$ &  \cellcolor{red!25}    $3.6 \times 10^{-4}$        \\
                     \hline
\multirow{2}{*}{G8}  & Median &  $97.85\%$ &  $93.51\%$& \cellcolor{blue!25}  $99.99\%$        & \cellcolor{blue!25}     $98.85\%$      & \cellcolor{blue!25}   $98.85\%$        &\cellcolor{blue!25}          $98.86\%$ & \cellcolor{red!25} $100\%$            \\
                     & MAD    &   $1.4 \times 10^{-2}$  & $8.9 \times 10^{-2}$& \cellcolor{blue!25}  $6.0 \times 10^{-3}$          &  \cellcolor{blue!25}    $1.0 \times 10^{-2}$       &  \cellcolor{blue!25} $1.0 \times 10^{-2}$          &   \cellcolor{blue!25}   $9.0 \times 10^{-3}$       & \cellcolor{red!25}    $5.9 \times 10^{-3}$          \\
                     \hline
\multirow{2}{*}{G9}  & Median &  $20.26\%$  &  $35\%$ & $81.62\%$           &  $80.55\%$          & $76.19\%$           &     \cellcolor{red!25} $97.95\%$       &  $81.24\%$            \\
                     & MAD    &  $2.0 \times 10^{-1}$  & $1.5\times 10^{-1}$ &  $1.7 \times 10^{-1}$          &  $2.3 \times 10^{-1}$          &  $2.5 \times 10^{-1}$          &         \cellcolor{red!25}  $1.6 \times 10^{-2}$  &  $5.6 \times 10^{-2}$            \\
                     \hline
\multirow{2}{*}{G19} & Median &  $99.89\%$  &  $99.68\%$  &   \cellcolor{blue!25} $99.92\%$         &   \cellcolor{blue!25} $99.91\%$        &            \cellcolor{blue!25} $99.92\%$&   \cellcolor{red!25}    $99.94\%$     &     \cellcolor{blue!25}     $99.91\%$     \\
				& MAD & $4.0 \times 10^{-4}$  &  $7.8 \times 10^{-4}$  &   \cellcolor{blue!25} $2.7 \times 10^{-4}$         &   \cellcolor{blue!25} $2.4 \times 10^{-4}$        &            \cellcolor{blue!25} $2.1 \times 10^{-4}$  &   \cellcolor{red!25}    $2.5 \times 10^{-4}$       &     \cellcolor{blue!25}     $2.6 \times 10^{-4}$   \\
                     \hline
\multirow{2}{*}{G24} & Median &  $99.59\%$ &  $59.75\%$  &\cellcolor{blue!25} $99.66\%$            &  \cellcolor{blue!25} $99.66\%$          &           \cellcolor{blue!25} $99.63\%$ &     $99.63\%$        &  \cellcolor{red!25}  $99.71\%$           \\
                     & MAD    &  $1.4 \times 10^{-3}$  & $7.4 \times 10^{-2}$ & \cellcolor{blue!25} $1.1 \times 10^{-3}$          & \cellcolor{blue!25} $6.1 \times 10^{-4}$          &   \cellcolor{blue!25} $8.5 \times 10^{-4}$        &            $1.8 \times 10^{-2}$&  \cellcolor{red!25} $3.9 \times 10^{-4}$      \\
                         \hline
\end{tabular}
\end{table}

The results show that the acquisition functions proposed in this paper
outperform naive LHS and the acquisition function  $\alpha_K$ proposed by Knudde \textit{et al.} \cite{knudde:active}. G9 has the worst median performance of $20.26\%$ for LHS,
where the volume of the feasible space is extremely small (about $0.5218\%$). Here too $\alpha_K$ performs worse than our acquisition function $\alpha_{PBE}$. 
The acquisition function $\alpha_E$ from Echard \textit{et al.} outperforms all other methods with a
median informedness of $97.95\%$. In \textbf{three out of the five problems},
$\alpha_{PBE}$ achieves the best median performance, while
$\alpha_E$ performs best in the rest of the problems. The best median for
any problem is at least $97.95\%$ with small MAD, demonstrating the
efficacy of the proposed and adapted methods over the current state-of-the-art.

\section{Conclusions}
\label{sec:concl}

This paper has examined the problem of feasible space identification for
computationally expensive problems. We have demonstrated an active learning
approach using Bayesian models (Bayesian search) and developed a range of
acquisition functions for this purpose. Our experiments show that our proposed acquisition function for Bayesian
search outperforms naive LHS, and the current state-of-the-art $\alpha_K$. 
%, achieving a median performance of at least
%$97.95\%$ across all problems. 
We propose that future work focusses on batch
 Bayesian search when it is possible to evaluate multiple solutions in
parallel.

%%%%%%%%%%%%%%%%%%5
\section*{Acknowledgements}

We acknowledge the support of the Supercomputing Wales project, which is
part-funded by the European Regional Development Fund (ERDF) via Welsh
Government.

% Bibliography
%\bibliographystyle{unsrt}
%\bibliography{references}
%{\small{\putbib}}
\putbib
\end{bibunit}

\pagebreak
\begin{bibunit}
\begin{center}
\begin{Large}	
\textbf{Supplementary Materials}
\end{Large}
\end{center}
\setcounter{section}{0}
\setcounter{equation}{0}

In the following supplementary sections, we present a range of related information that the readers may find useful.

\section{Related Work in Reliability Engineering}
\label{sec:rel}

The reliability engineering literature has much work devoted to
system reliability analysis (SRA). SRA is applied when there are multiple failure modes in a
system \cite{bichon:sra}, and Yang \textit{et al.} \cite{yang:system} provide a comprehensive review of work
in this area. In these cases, a sequential search approach is adopted to
constructing constraint models, which are then used to compute the probability
of failure. Here, their ultimate goal is to estimate the total volume of the infeasible
space or the excursion set \cite{vazquez:pof}.

The earliest approaches to modelling the boundary of the feasible space used
either polynomials (typically first or second-order) \cite{freudenthal:safety,
  rackwitz:structural, castillo:uncertainty} or support vector machines (SVMs)
\cite{hurtado:structural}. However, these approaches are limited.
Under multiple constraints, the boundary is often highly-non-linear, and may 
even be discontinuous \cite{perrin:active}.

Polynomials and SVMs therefore perform poorly in modelling the boundary
directly. To solve this problem, others have attempted to model the constraint
functions instead. Attempts have been made using neural networks
\cite{papadopoulos:accelerated} and SVMs \cite{bourinet:assessing}, but since these
methods only produce point-predictions, there is no
quantification of uncertainty. The predictions of the feasible space may therefore be
misleading \cite{cadini:improved}.

Recently, \mGP models have shown promise as a framework for active
learning \cite{bichon:sra, echard:ak, fauriat:ak, hu:efficient, wang:radial,
  yang:system}.
The \mGP approach is similar to Bayesian search. The difference is that
 the \mGP approach maximises the acquisition function using a
variant of Monte Carlo search to find the next promising sample to simulate.
This is an important distinction, since the Monte Carlo
search does not perform as well as evolutionary search methods.
The Bi-population Covariance Matrix Adaptation Evolutionary Strategy
(Bi-POP-CMA-ES) has been shown to perform better than Monte
Carlo\cite{hansen:compare}, so we propose this method for maximising the infill
function.

Many of the popular approaches adopt a composite criterion approach
\cite{yang:system}. In these approaches, an acquisition function is created with
the aim of improving the estimation of each relevant constraint. Each model is
selected based on the mean predictions. These predictions determine which
acquisition function should be used to select the parameters for the next
expensive simulation. This approach is effective, but there are some drawbacks
which mean that they are not suitable for our approach. Firstly, a model is
selected by considering all Monte Carlo samples. The adaptation in our framework therefore required a reformulation of the combined acquisition function. Secondly, irrespective of the
reformulation, the selection of the model requires reliance on the mean
predictions. This may be misleading, particularly during the early stages of the
search where data is sparse. Finally, the composite criterion approach tends to
underperform if the constraint functions have a difference in scales and cannot be
easily normalised \cite{yang:system}. Our proposed acquisition function does not
require the model selection step. Instead, it combines predictive distributions
from all models. This allows the computation of the utility of a candidate
solution using the models without the need to normalise the value of individual constraints.

\section{Modelling Scalarised Constraints}

We can use a scalarisation approach as an alternative method for dealing with
the multiplicity of constraints. This approach could encapsulate all constraints
into a single function so that any violation of the scalarised constraint is
equivalent to infeasibility \cite{li:maxform}:
 \begin{align}
 \label{eq:maxform}
s(\bx) = \max_{l=1}^L \left(g_l (\bx)   - t_l\right) \leq  0.
 \end{align}

Here, the response of $s:\Real^{n} \rightarrow \Real$ is only greater than $0$
for a design vector resulting in an infeasible solution, iff at least one
of the component $l$th constraints is violated ($g_l(\bx) > t_l$).
It is possible to construct a model of this scalarisation instead of constructing one for each constraint. 
From a reliability engineering perspective, such \emph{mono-surrogate} approaches
 are known to be inferior \cite{yang:system}. We confirmed this to be
true by a small test. Hence, we excluded the approach from our investigation.

\section{Acquisition Function by Knudde \textit{et al.}}

In this section, we provide a simplification of the acquisition function proposed in \cite{knudde:active}. The general form of the acquisition function is given as: 
\begin{align}
\label{eq:knudde-orig}
\alpha_{K[l]}(\bx) = &3 \entropy[\prob (g_l \Given \bx, \mD_l)]\\\nonumber 
&- \entropy[p(g_l \Given \bx, \mD_l, g_l > b)] \\\nonumber 
&- \entropy[p(g_l \Given \bx, \mD_l, a < g_l < b)] \\\nonumber 
&- \entropy[p(g_l \Given \bx, \mD_l, g_l < a)],
\end{align}
where, the feasible space is defined to be in the range $a \leq g_l (\bx) \leq b$.

To transform it into the form $g_l \leq t_l$, we set $a = -\infty$, and derive a simpler representation. 

Firstly, if $a = -\infty$, then $\entropy[p(g_l \Given \bx, \mD_l, g_l < a)] = 0$ and $\entropy[p(g_l \Given \bx, \mD_l, a < g_l < b)] = \entropy[p(g_l \Given \bx, \mD_l, g_l < b)]$. 

Like Knudde \textit{et al.}, we note that the second and third terms in equation \eqref{eq:knudde-orig} are entropies of truncated Gaussian. The general form of entropy of a truncated Gaussian is given by:
\begin{align}
\entropy[\prob(f \Given u \leq f \leq v)] = \ln(\sqrt{2\pi e}\sigma_f Z) + \frac{1}{2Z} [\Psi \Phi(\Psi) - \Omega \Phi(\Omega)],
\end{align}
where $u$ and $v$ are constants, $Z = \Phi(\Omega) - \Phi(\Psi)$, $\Psi = \frac{u - \mu_f}{\sigma_f}$ and $\Omega = \frac{v - \mu_f}{\sigma_f}$.

Using this, the second and third terms become: 
\begin{align}
&\entropy[p(g_l \Given \bx, \mD_l, g_l < b)] = \ln(\sqrt{2\pi e}\sigma_l \Phi(\tau_l)) - \frac{1}{2\Phi(\tau_l)}\tau_l \Phi(\tau_l),\\
&\entropy[p(g_l \Given \bx, \mD_l, g_l > b)] = \ln(\sqrt{2\pi e}\sigma_l (1-\Phi(\tau_l))) + \frac{1}{2(1-\Phi(\tau_l))}\tau_l \Phi(\tau_l),
\end{align}
where, $\tau_l = \frac{t_l -\mu_l}{\sigma_l}$.

Replacing these terms in \eqref{eq:knudde} and using logarithmic identities, we derive:
\begin{align}
\alpha_{K[l]}(\bx) = \frac{1}{2} \ln(2\pi e \sigma_l^2) - \ln(~\Phi(\tau_l)~(1-\Phi(\tau_l))~).
\end{align}

\section{Additional Results}

\begin{figure}[!t]
    \centering
    \begin{subfigure}[b]{0.45\textwidth}
        \includegraphics[scale=0.475, trim={5mm 16mm 0 17mm}, clip=true]{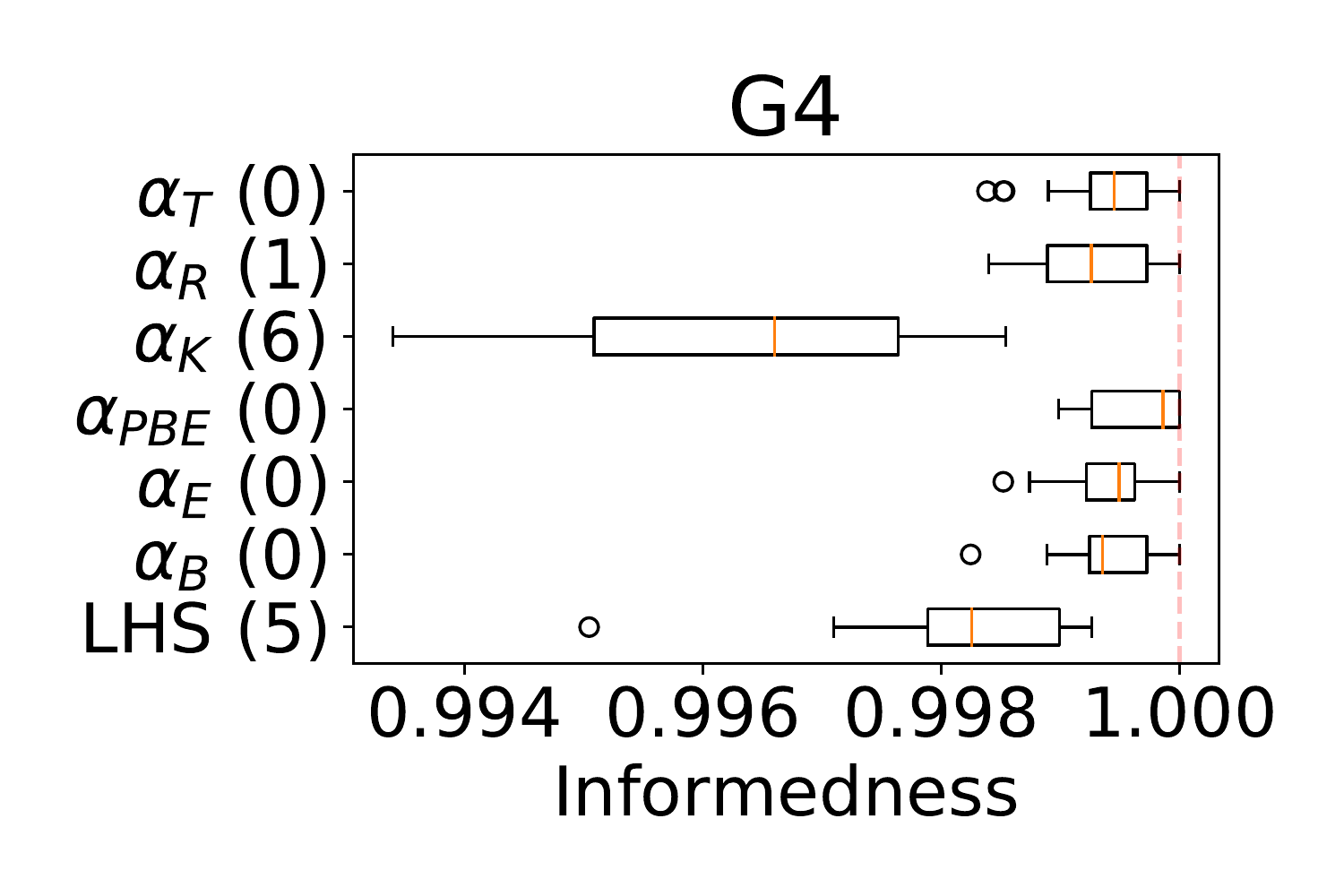}\caption{G4}
        \label{fig:g4}
    \end{subfigure}
    \hfill%add desired spacing between images, e. g. ~, \quad, \qquad, \hfill etc. 
    \begin{subfigure}[b]{0.45\textwidth}
        \includegraphics[scale=0.475, trim={25mm 16mm 0 17mm}, clip=true]{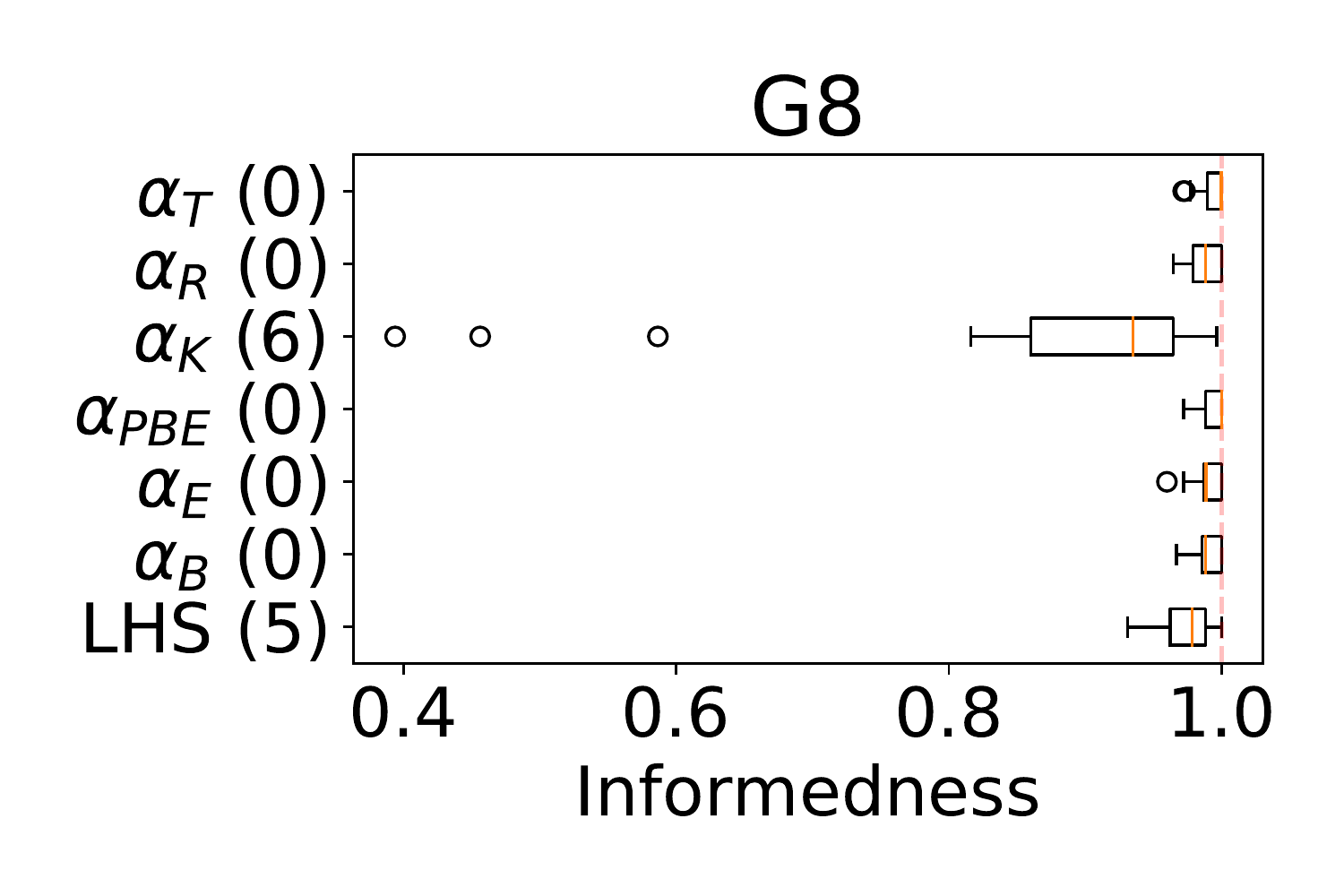}\caption{G8}
        \label{fig:g8}
    \end{subfigure}
    
        \begin{subfigure}[b]{0.45\textwidth}
        \includegraphics[scale=0.475, trim={5mm 16mm 0 17mm}, clip=true]{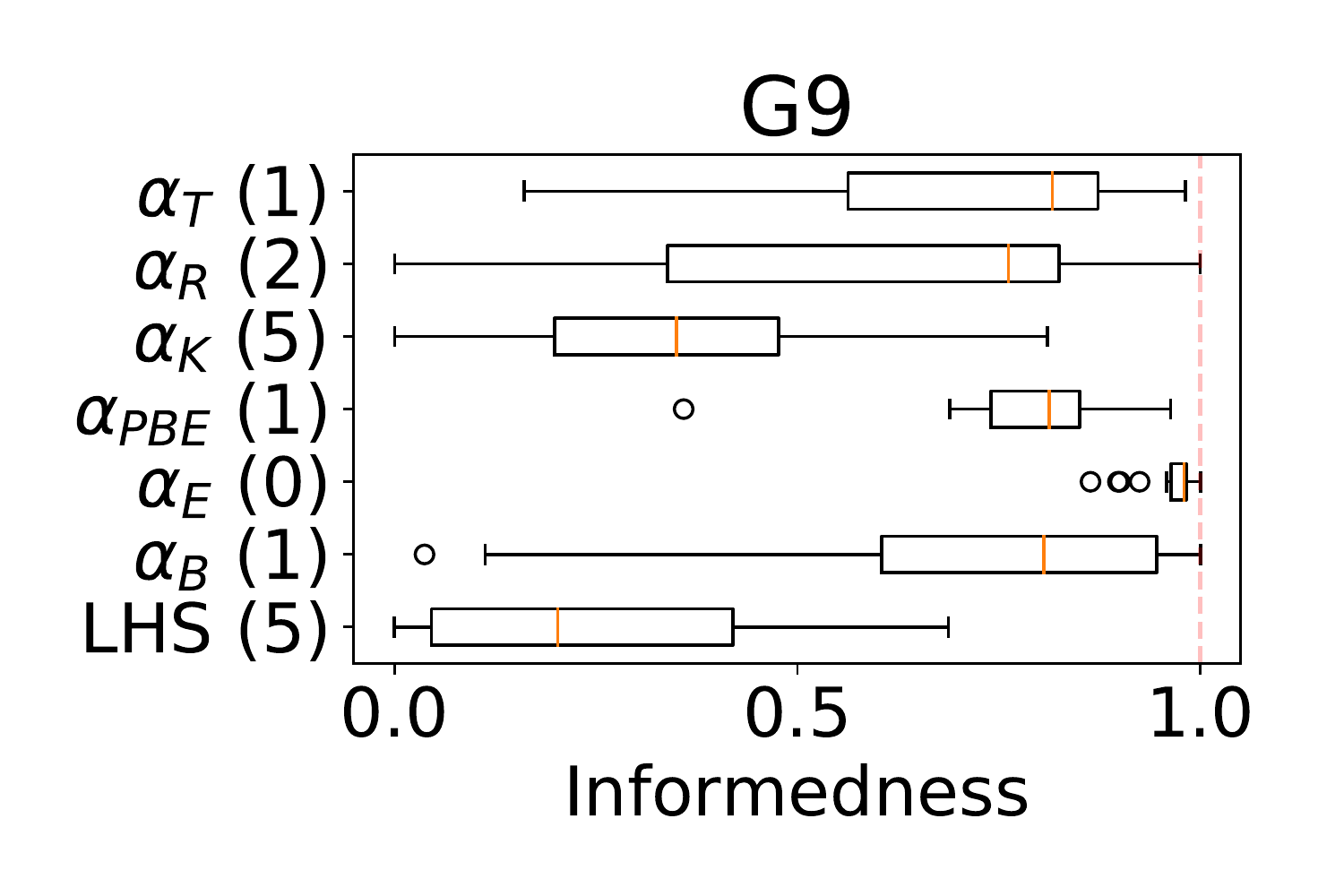}\caption{G9}
        \label{fig:g9}
    \end{subfigure}
\hfill%add desired spacing between images, e. g. ~, \quad, \qquad, \hfill etc. 
    \begin{subfigure}[b]{0.45\textwidth}
        \includegraphics[scale=0.475, trim={25mm 16mm 0 17mm}, clip=true]{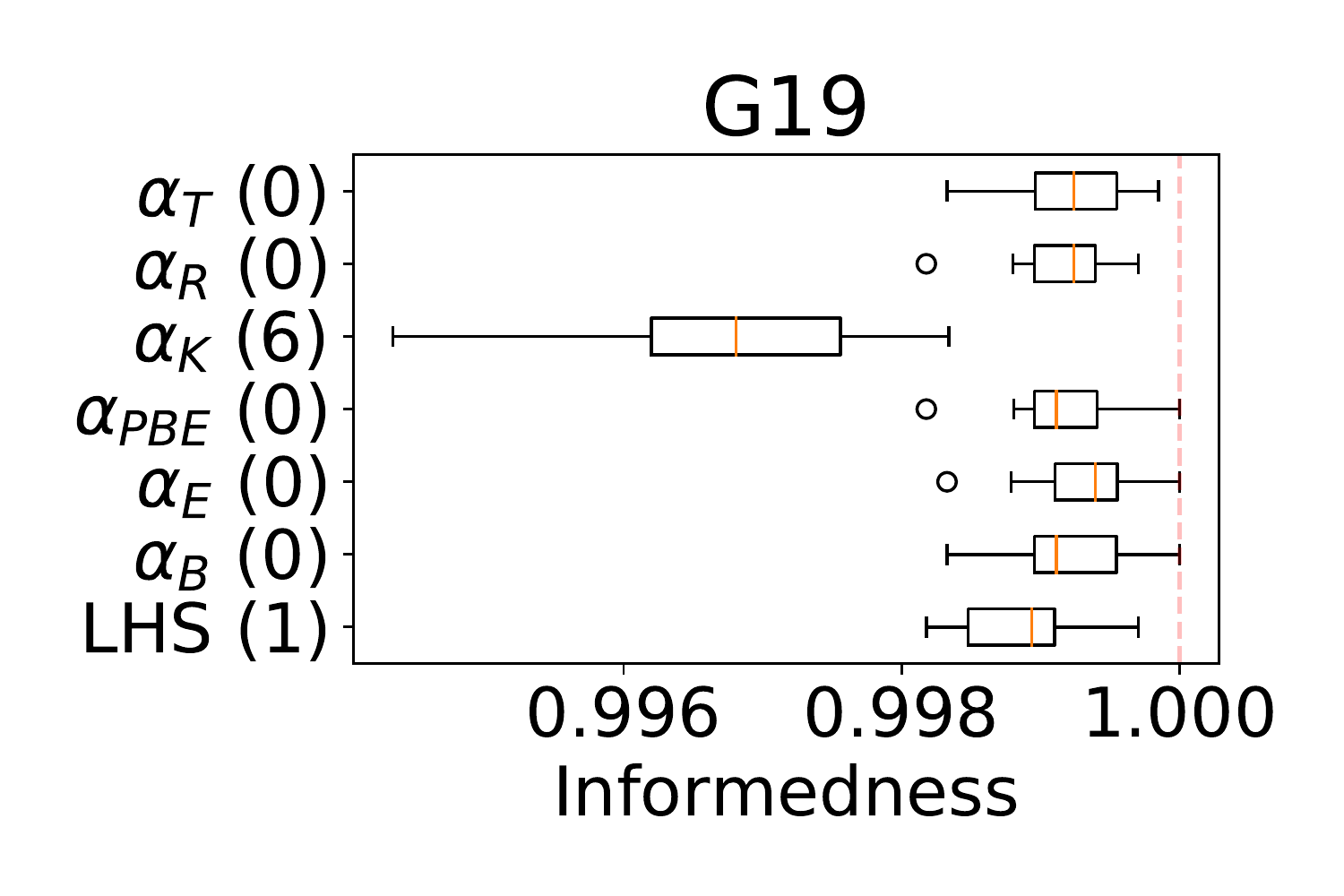}\caption{G19}
        \label{fig:g19}
    \end{subfigure}

%    \begin{subfigure}[b]{0.45\textwidth}
%        \includegraphics[scale=0.5, trim={5mm 8mm 0 17mm}, clip=true]{G9_informedness.pdf}\caption{G9}
%        \label{fig:g9}
%    \end{subfigure}
%    \hfill%add desired spacing between images, e. g. ~, \quad, \qquad, \hfill etc. 
%      %(or a blank line to force the subfigure onto a new line)
%    \begin{subfigure}[b]{0.45\textwidth}
%        \includegraphics[scale=0.5, trim={25mm 8mm 0 17mm}, clip=true]{G19_informedness.pdf}\caption{G19}
%        \label{fig:g19}
%    \end{subfigure}

    \begin{subfigure}[b]{0.45\textwidth}
        \includegraphics[scale=0.475, trim={5mm 8mm 0 17mm}, clip=true]{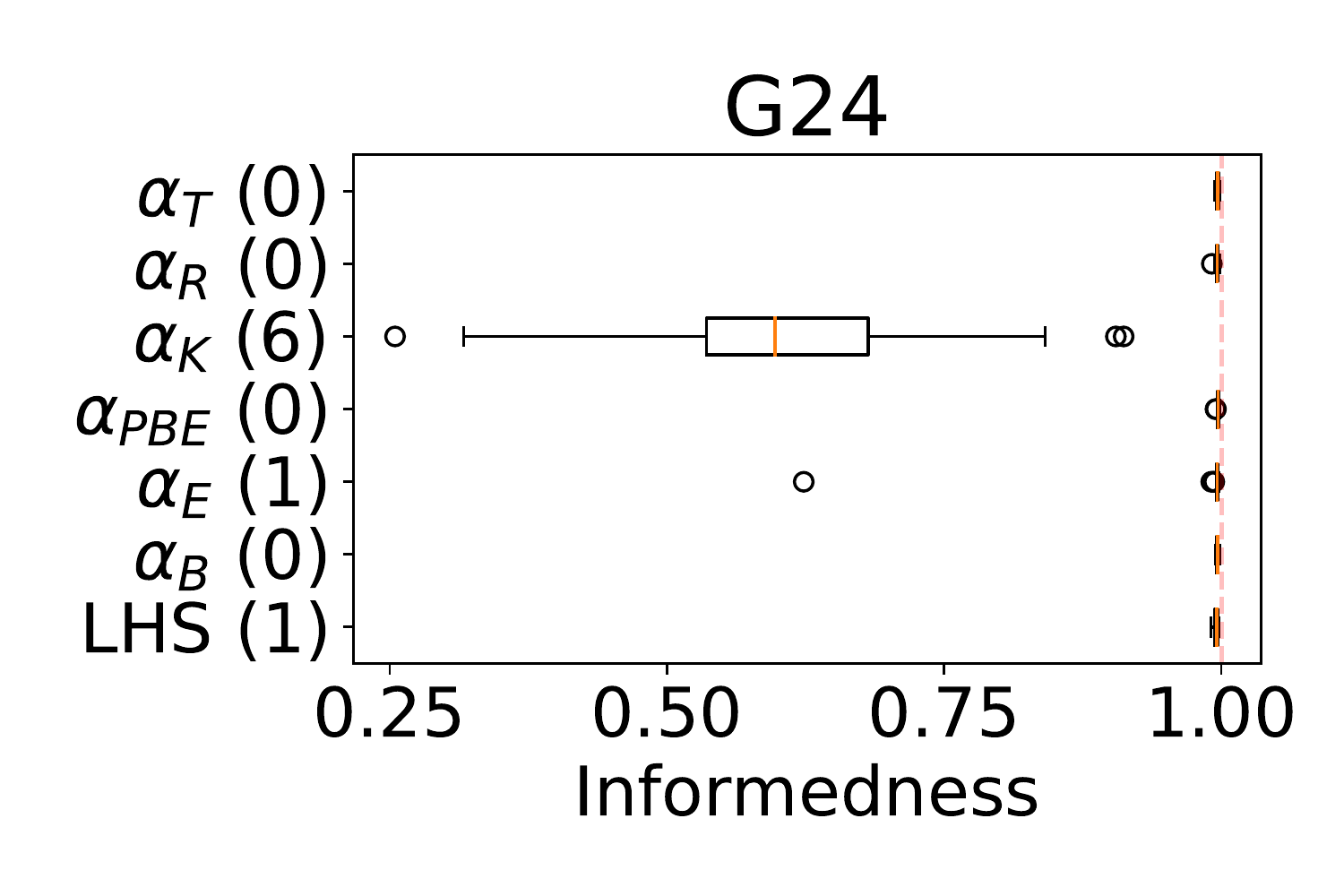}\caption{G24}
        \label{fig:g24}
    \end{subfigure}
    \hfill %add desired spacing between images, e. g. ~, \quad, \qquad, \hfill etc. 

\caption{Performance comparison for the selected test problems. The number in brackets show how many alternative acquisition functions beats it statistically significantly in informedness. The state of the art $\alpha_K$ is statistically significantly beaten in all problems by the best, and in G4 it is beaten by the naive LHS. }
\end{figure}

Here we show the full results of final performances through box plots in Figures \ref{fig:g4} to \ref{fig:g24}. Clearly, the state-of-the-art $\alpha_K$ is worse than the proposed methods, and sometimes it is even worse than a naive LHS. 

\begin{figure}[!t]
    \centering
    \begin{subfigure}[b]{0.45\textwidth}
        \includegraphics[scale=0.475, trim={3mm 10mm 0 17mm}, clip=true]{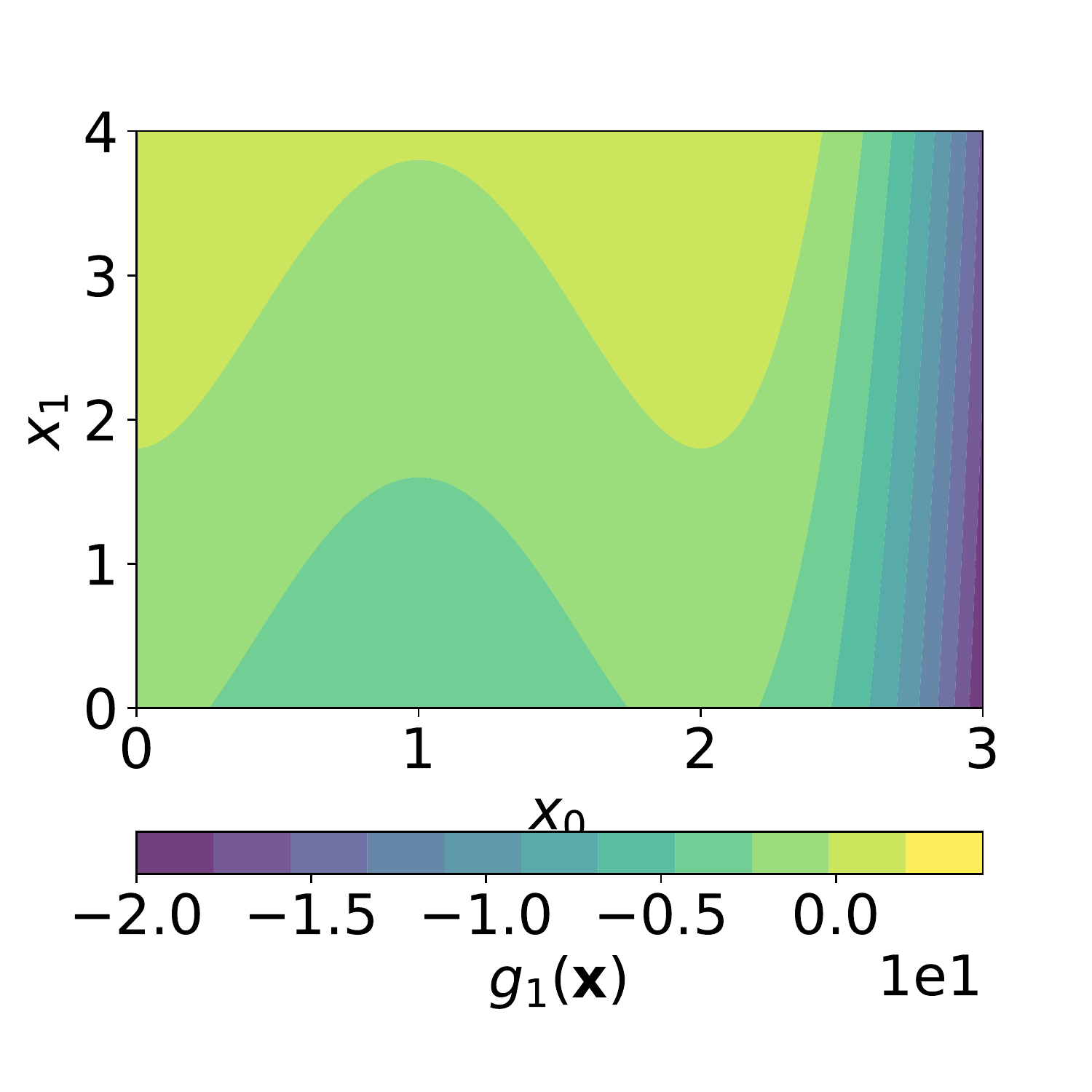}%\caption{G4}
        \label{fig:g24-1}
    \end{subfigure}
    \hfill%add desired spacing between images, e. g. ~, \quad, \qquad, \hfill etc. 
    \begin{subfigure}[b]{0.45\textwidth}
        \includegraphics[scale=0.475, trim={25mm 10mm 0 17mm}, clip=true]{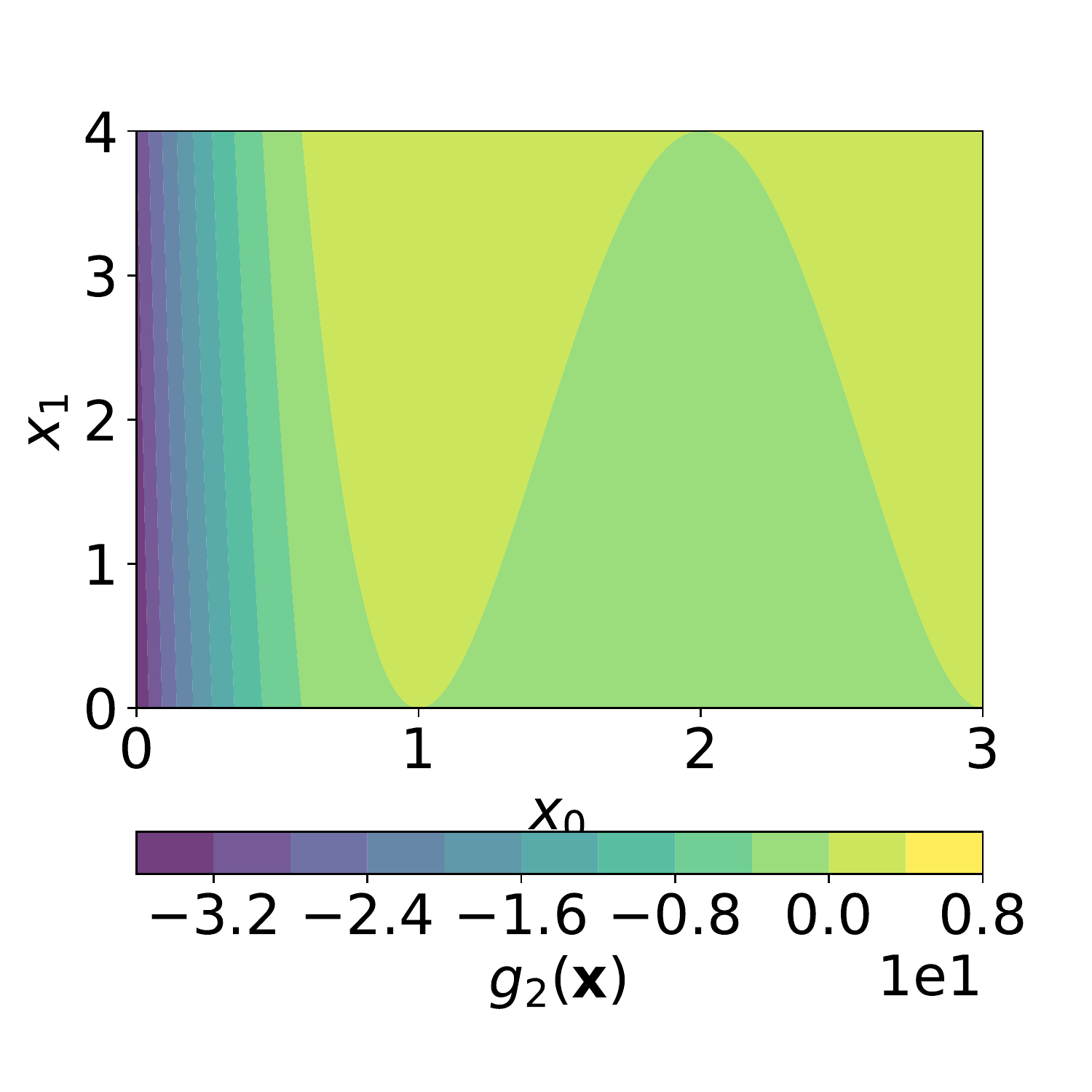}%\caption{G8}
        \label{fig:g24-2}
    \end{subfigure}
    
        \begin{subfigure}[b]{0.45\textwidth}
        \includegraphics[scale=0.475, trim={3mm 10mm 0 17mm}, clip=true]{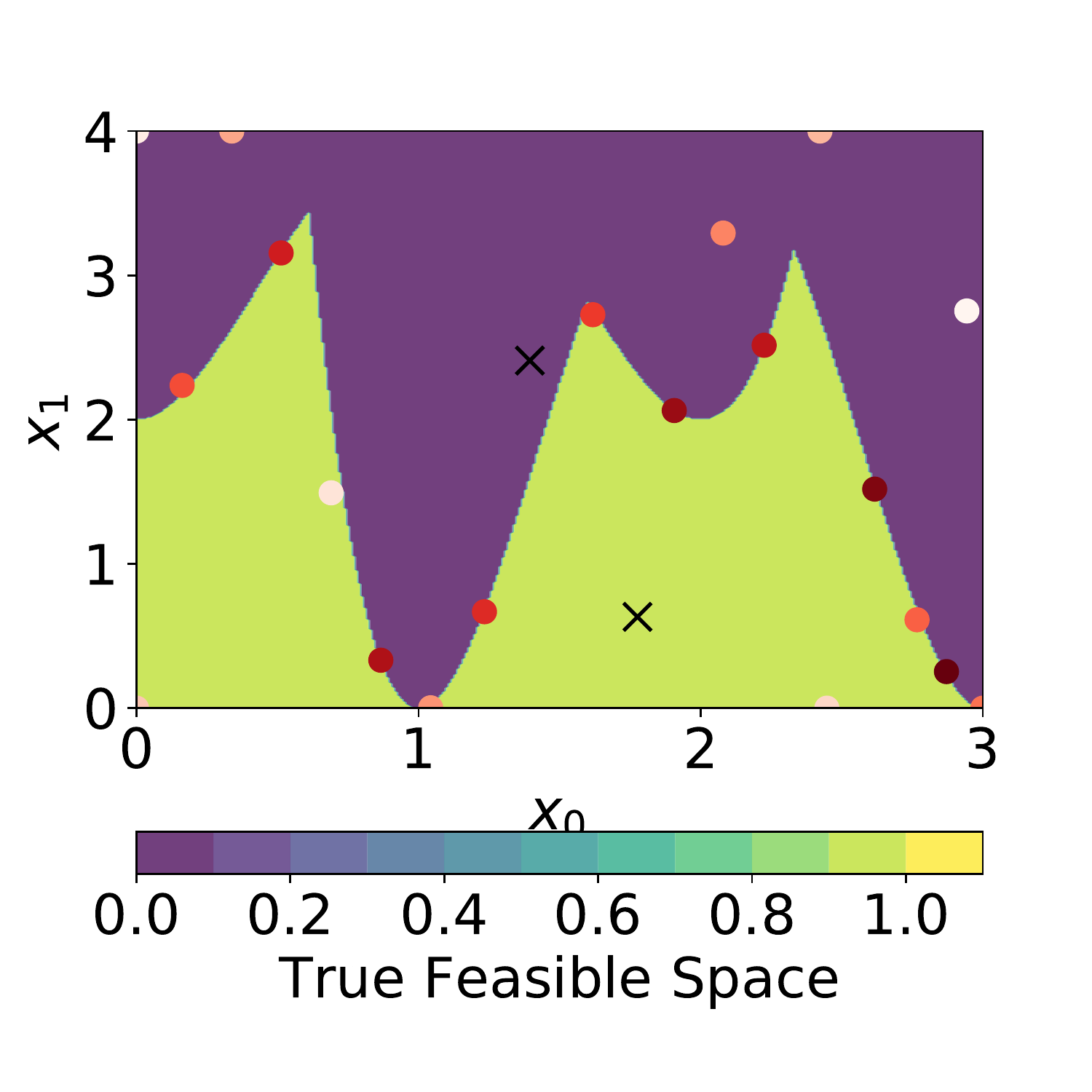}%\caption{G9}
        \label{fig:g24-tf}
    \end{subfigure}
\hfill%add desired spacing between images, e. g. ~, \quad, \qquad, \hfill etc. 
    \begin{subfigure}[b]{0.45\textwidth}
        \includegraphics[scale=0.475, trim={25mm 10mm 0 17mm}, clip=true]{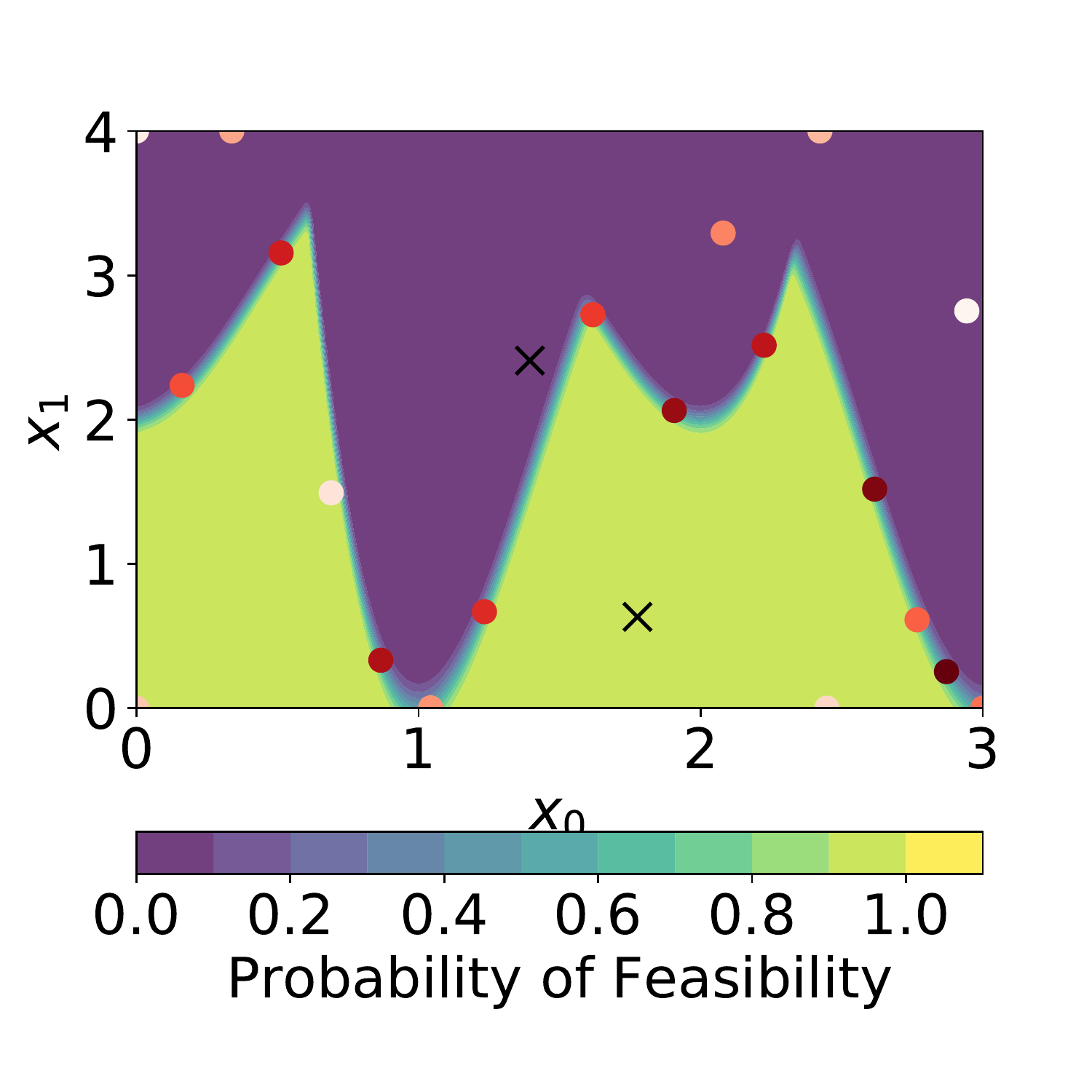}%\caption{G19}
        \label{fig:g24-ef}
    \end{subfigure}
    
    \caption{Illustration of the classifier performance in G24. The top row depicts the function landscape for the two constraint functions $g_1(\bx)$ and $g_2(\bx)$. On the bottom row, the black crosses portray the initial samples, and the coloured dots show the samples taken using the proposed acquisition function with darker colour representing later samples. On the bottom left, we show the true feasible space with light green showing the feasible space. On the bottom right, we show the predicted feasible space using our method after only $22$ samples were taken. This clearly shows that the samples taken are often on the boundary, and the final feasibility predictions are good estimations of the true feasible space.  \label{fig:contours}}
    \end{figure}

We also illustrate an example classifier output using the proposed acquisition function for G24 in Figure \ref{fig:contours}.

\section{Relationship with Constrained Optimisation}

Apart from design exploration applications, the ability to accurately determine the
feasible space using our method may also be useful in constrained
global optimisation of problems. We believe that our work can be complementary to
constrained optimisation approaches, since we can construct a classifier of
feasibility before performing optimisation. This approach may
be useful during the preliminary stages of optimisation when
we are trying to understand the problem at hand, and may assist us in
determining the true constraints and the character of objective functions.

In Bayesian optimisation for constrained problems, most acquisition functions
locate solutions with certain characteristics: those with a high
probability of feasibility and those with the greatest expected improvement. For
more on this subject, we refer the interested reader to
\cite{picheny:sur-opt,li:maxform,bagheri:constraint}.

The acquisition functions proposed in this paper do not directly target optimisation.
Instead, they seek to locate solutions \textit{at the
  boundary between the feasible and infeasible spaces}. In most cases, this will
improve the accuracy of the feasibility classification.
However, if the acquisition function is driven by the probability
of feasibility only, the algorithm is biased to choose solutions
away from the boundary. In fact, Knudde \textit{et al.} \cite{knudde:active}
compared against probability of feasibility as a basis of search to improve the
classifier. Their results convincingly showed that probability of feasibility
understandably performs worse than their proposed method. As such, we refrained
from comparing against probability of feasibility in this paper.

  In future, we propose to investigate the efficacy of creating an accurate
  classifier for feasibility prior to optimisation, and evaluating its
  usefulness in constrained optimisation.

\begin{small}
%%%%%%%%%%%%%
\putbib
\end{small}
\end{bibunit}

\end{document}